\documentclass[journal]{IEEEtran}
\usepackage{amsmath}
\usepackage[utf8]{inputenc} 
\usepackage[T1]{fontenc}    
\usepackage{hyperref}       
\usepackage{url}            
\usepackage{booktabs}       
\usepackage{amsfonts}       
\usepackage{nicefrac}       
\usepackage{microtype}      
\usepackage{cleveref}       
\usepackage{lipsum}         
\usepackage{graphicx}
\usepackage{natbib}
\usepackage{doi}
\usepackage{array}
\usepackage{caption}
\setcitestyle{numbers,square}
\ifCLASSINFOpdf
\else
\fi
\begin{document}
%
\title{Vision-assisted Avocado Harvesting with Aerial Bimanual Manipulation}
%
%
%

\author{Zhichao~Liu \textsuperscript{\textdagger},
        Jingzong~Zhou \textsuperscript{\textdagger},
        Caio~Mucchiani,
        and~Konstantinos~Karydis*,
\thanks{Authors are all from Dept. of Electrical and Computer Engineering
	University of California, Riverside}
\thanks{\textsuperscript{\textdagger} Equal contribution}
\thanks{* kkarydis@ece.ucr.edu}
}

%
%

\markboth{}%
{Shell \MakeLowercase{\textit{et al.}}: Bare Demo of IEEEtran.cls for IEEE Journals}
%



\maketitle

\begin{abstract}
Robotic fruit harvesting holds potential in precision agriculture to improve harvesting efficiency. While ground mobile robots are mostly employed in fruit harvesting, certain crops, like avocado trees, cannot be harvested efficiently from the ground alone. This is because of unstructured ground and planting arrangement and high-to-reach fruits. In such cases, aerial robots integrated with manipulation capabilities can pave new ways in robotic harvesting. This paper outlines the design and implementation of a bimanual UAV that employs visual perception and learning to autonomously detect avocados, reach, and harvest them. The dual-arm system comprises a gripper and a fixer arm, to address a key challenge when harvesting avocados: once grasped, a rotational motion is the most efficient way to detach the avocado from the peduncle; however, the peduncle may store elastic energy preventing the avocado from being harvested. The fixer arm aims to stabilize the peduncle, allowing the gripper arm to harvest. The integrated visual perception process enables the detection of avocados and the determination of their pose; the latter is then used to determine target points for a bimanual manipulation planner. Several experiments are conducted to assess the efficacy of each component, and integrated experiments assess the effectiveness of the system. 
\end{abstract}


%
\IEEEpeerreviewmaketitle

\section{Introduction}\label{sec:intro}

Robotics have been increasingly integrated into precision agriculture~\cite{sparrow2021robots} as they can help address some of the current challenges associated with a lack of agricultural workforce~\cite{gonzalez2020automating}, take over strenuous activities for workers~\cite{racine2012farming}, improve utilization of resources~\cite{bagagiolo2022greenhouse}, and enhance productivity in repetitive tasks~\cite{kulkarni2020applications}.   
To date, ground mobile robots have been mostly deployed to support tasks such as in-field transportation and harvesting~\cite{grieve2019challenges}. 
Yet, not all cases may be amenable to ground mobile robot-based fruit harvesting in tree crops~\cite{vougioukas2019agricultural}, like avocados (which is the focus of this work). 
There are cases where the ground morphology (e.g., not equally-spaced rows) and the planting arrangement (e.g., not growing along a trellis) may be less structured or the fruits to be at a height where deploying
a ground-based solution (e.g., a telescopic mechanism to reach them) would be too
costly to develop, if possible at all. 
In such cases, aerial manipulators (i.e. Unmanned Aerial Vehicles [UAVs] integrated with manipulation capabilities) may offer a solution.

While most UAVs in precision agriculture have focused on remote and proximal sensing~\cite{raeva2019monitoring,bertalan2022uav,awais2022uav} and spraying~\cite{hafeez2023implementation}, a growing research direction has been emerging where these robots can perform some physical interactions with the crop. 
Examples of aerial manipulation in agriculture include leaf retrieval~\cite{orol2017aerial}, cargo transfer~\cite{kotarski2022toward}, health monitoring by vibration~\cite{gonzalez2024controlled}, pollination~\cite{pandey2022towards}, and collection of endangered cliff plants~\cite{la2022collecting}. 
Fruit harvesting using UAVs and a single degree-of-freedom (DoF) manipulator was also explored~\cite{park2024fruit}. 
In addition, a notable commercial effort includes Tevel~\cite{tevel} which developed and successfully deployed a tethered multi-UAV system for fruit harvesting. 

Despite this progress, UAVs remain underutilized in physical manipulation tasks in agriculture, primarily because of their inherent physical constraints and the complexity of the overall control system. An additional challenge that underlines practical deployment is the integration of perception for task execution (e.g., fruit detection and localization for the robot to reach), which adds another layer of integrated mechanical and computational complexity to be addressed. 
Existing research efforts have attempted to improve the efficiency of aerial manipulation as well as explore its deployment in the context of different applications. 
Notable examples of achieving efficient aerial manipulation include well-designed end-effectors such as multi-fingered grippers~\cite{pounds2011yale}, compliant grippers~\cite{liu2022safely}, and multi-shaped grippers~\cite{orol2017aerial}. 
Other designs include a planar translational driving system for pull and push forces~\cite{miyazaki2021development}, and a hybrid rigid-compliant manipulator deployed onto a hexarotor UAV~\cite{suarez2021cartesian}. 
Technological advancements further enhance efficiency, such as learning-based planning and control~\cite{kim2018cooperation}, VR-assisted teleoperation~\cite{kocer2022immersive}, and tactile feedback~\cite{yashin2019aerovr}.
Aerial manipulation has been widely applied in various fields such as object pickup~\cite{pounds2011grasping}, structure assembly~\cite{trujillo2019structure}, door opening~\cite{lee2020aerial}, intercepting flying targets~\cite{liu2022safely}, installing bird diverters \cite{suarez2021experimental2}, inspecting infrastructure \cite{suarez2020aerial}, and powerline maintenance \cite{suarez2021experimental,  nekoo2023constrained}.

However, in several cases, crop harvesting may necessitate a bimanual manipulation solution. 
An example of interest to this work is harvesting avocados: once grasped, a rotational motion is the most effective way to detach the avocado from the peduncle without cutting it~\cite{zhou2024design}; however, the peduncle may store significant elastic energy which may prevent the avocado from being harvested efficiently. 
Some works have addressed bimanual harvesting for harvesting aubergines~\cite{sepulveda2020robotic}, kiwifruit~\cite{he2022double}, and pears~\cite{yoshida2022automated}, albeit with ground mobile robots. 
Developing bimanual aerial manipulators, instead, faces significant challenges because of the limited payload capacity and available mounting area, as well as energy utilization and the resulting operational time~\cite{suarez2020compliant}. 
Previous research has introduced bimanual aerial manipulators~\cite{suarez2020compliant,suarez2017lightweight, grau2018design, suarez2018design, suarez2019compliant,  perez2020hecatonquiros} and deployed them for pickup~\cite{korpela2012mm, suarez2018design, grau2018design}, pipe structures inspection~\cite{suarez2020compliant} and multi-shaped objects handling~\cite{perez2020hecatonquiros}. 
However, current bimanual aerial robots lack the capacity for autonomous fruit (specifically avocado) harvesting because of the absence of distinct arms with task-appropriate end-effectors as well as the lack of integration of a comprehensive perception pipeline for onboard fruit detection and pose estimation. 

In this work, we develop a bimanual aerial manipulator with integrated visual perception and learning capabilities and deploy it for autonomous avocado harvesting. 
The robot employs two distinct arms, each designed to perform different yet complementary tasks, which, when combined, can enable efficient avocado harvesting. 
One arm is tasked with grasping and holding the peduncle in place, while the other arm grasps the avocado and applies a rotational motion to detach it. 
In our previous work~\cite{zhou2024design}, we have shown that such a harvesting motion is effective, and we have developed an appropriate end-effector mounted onto a commercial mobile manipulator. 
Herein, we improve upon the design of that end-effector to make it amenable for aerial bimanual manipulation along with designing the complete arm that this end-effector is mounted on. 
Visual perception and learning are used to enable the autonomous detection of avocados. 
This information is then used to determine their 3D pose, which in turn is used as input to a motion planner for the dual-arm system (i.e. goal pose for each arm). 
The overall system can thus autonomously detect avocados, reach them, and harvest them. 
A series of different experiments are conducted to assess the efficacy of each component and integrated experiments evaluate the effectiveness of the overall developed system in controlled settings. 
In all, this work offers a complete pipeline for developing and deploying bimanual aerial robots in practical applications and lays the foundation for field deployment and study involving completely autonomous robotic avocado harvesting.

In the remainder of this article we first present a detailed description of the design of our dual-arm system and the deployed end-effectors and derive the corresponding kinematics analysis (Section~\ref{sec:two}). 
We then detail how to enable the system with autonomous capabilities, specifically underlying visual detection and learning for avocado detection and pose estimation, and bimanual manipulation planning (Section~\ref{sec:three}). 
Results from evaluating each component as well as the overall system are provided and discussed in Section~\ref{sec:four}, while in Section~\ref{sec:five} we summarize our findings and discuss potential extensions and improvements. 

\section{Design and Analysis of the Bimanual Aerial Robot}\label{sec:two}
\subsection{Overall Design and Major Components}
\begin{figure*}[ht]
\vspace{-12pt}
    \centering
    \includegraphics[width=0.99\textwidth]{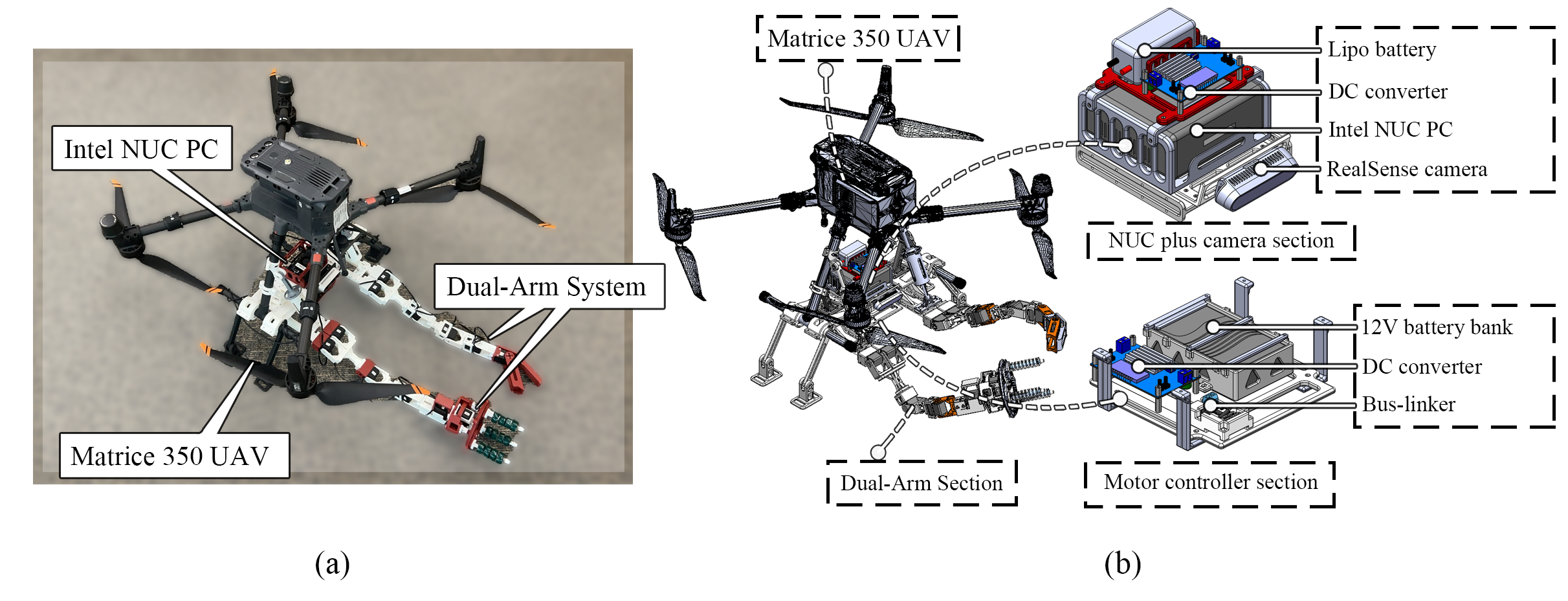}
    \vspace{-12pt}
    \caption{(a) The prototype bimanual aerial robot developed in this work. (b) Detailed view of key individual components.}
    \label{fig:uav_components}
    \vspace{-10pt}
\end{figure*}

The physical prototype of our developed bimanual aerial robot is shown in Figure \ref{fig:uav_components}-(a), while a more detailed view of different components is provided in Figure~\ref{fig:uav_components}-(b). 
The overall system is built upon a commercial quadrotor UAV (DJI Matrice 350). The UAV is at its standard configuration as provided by the manufacturer and uses two OEM batteries. 
We have retrofitted the UAV with a custom-built dual-arm system, attached below the UAV using a custom-built 3D-printed mounting base. 
The specific choice was made to address propeller downwash that could create significant vibrations to the arms. 
An onboard computer (Intel NUC with an i7-10710U processor, 16 GB RAM, and 512 GB SSD storage) is used for dual-arm motion planning and control as well as vision-based avocado detection and localization). 
A stereo camera (Intel RealSense D435i) is used for scene understanding, whereas a motor controller assembly is responsible for low-level control of the arms' motors (LX-16A bus servomotors). 
The onboard computer and camera are powered by a 22.2V LiPo battery; the motor controller is powered by a separate 12V lithium-ion power bank. 
DC converters are used to provide the appropriate voltage (19V for mini PC and 7V for servomotors). 
A bus-linker processes control messages from the onboard computer to the servomotors. 
All added components required for the dual-arm assembly are organized in a way that the total robot's center of mass (CoM) remains as close to the UAV's original CoM. 
The total weight of our bimanual aerial robot including batteries is 11.4 kg. 
The weight of the dual-arm system is 5.63 kg. 

\subsection{Design and Kinematic Analysis of the Dual-arm System}

\begin{figure*}[ht]
\vspace{0pt}
    \centering
    \includegraphics[width=0.78\textwidth]{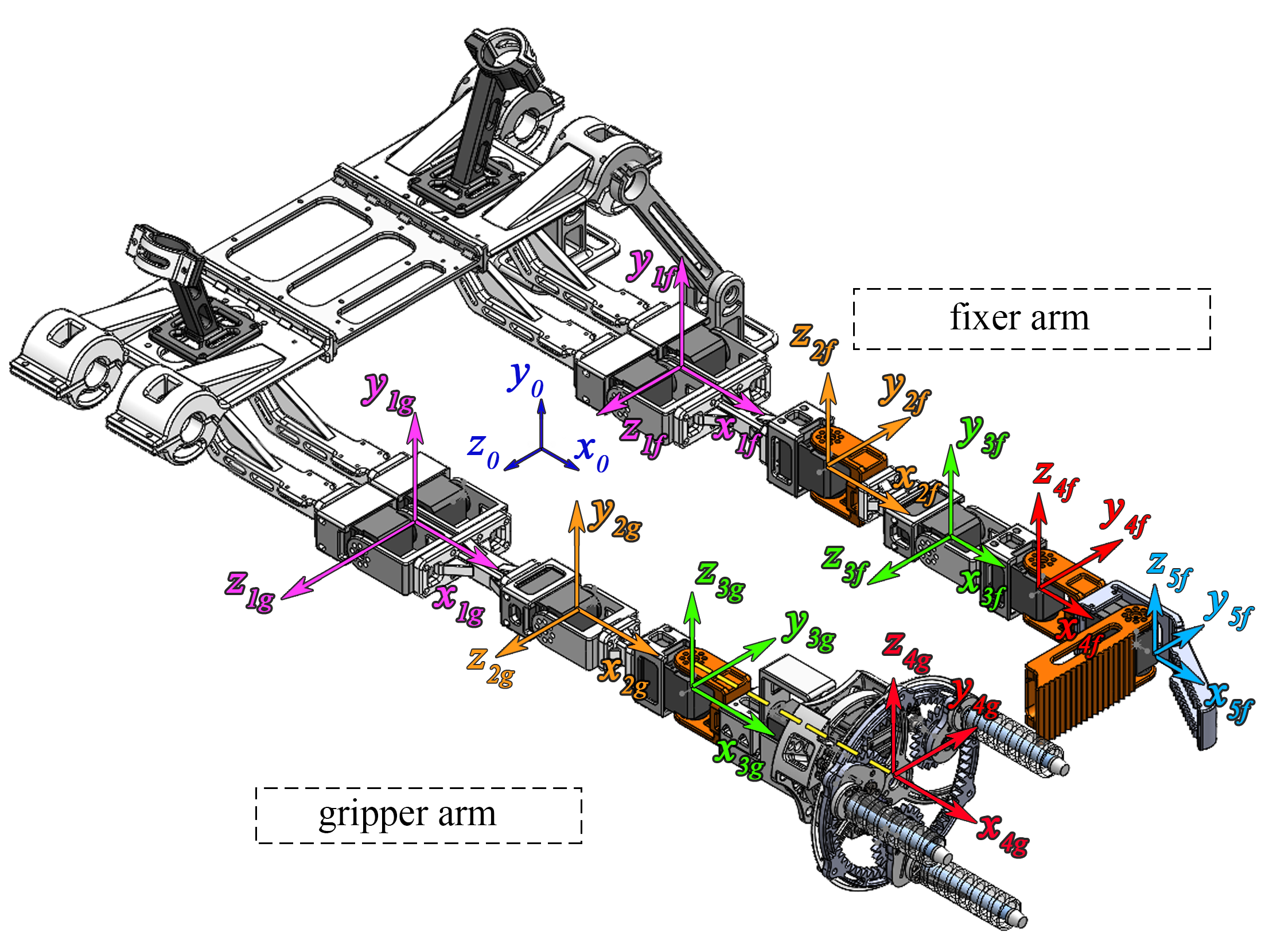}
    \vspace{-12pt}
    \caption{Detailed Dual-arm Assembly and Associated Coordinate System Frames.}
    \label{fig:dh_two_arm}
    \vspace{-12pt}
\end{figure*}

The dual-arm assembly (Figure~\ref{fig:dh_two_arm}) is designed to be lightweight. 
It is primarily 3D printed (Bambu Lab X1-Carbon Combo 3D printer) using carbon-fiber-reinforced material (PAHT-CF) for critical components and basic PLA material for non-critical components. 
The assembly comprises a ``fixer" arm and a ``gripper" arm, each with a distinctive yet complementary role. 
The role of the former is to hold the peduncle of the avocado, whereas the latter aims to engage with the avocado and harvest it. 
The fixer arm has four DoFs (excluding the fixer end-effector), while the gripper arm has three DoFs (also excluding the gripper end-effector). 
All joints in both arms (excluding the two end-effectors) are revolute and are powered by either one or two (see frames \emph{1g} and \emph{1f} in Figure~\ref{fig:dh_two_arm}) servomotors. 
At the operating voltage of 7V, the LX-16A bus servo delivers a torque of about 17 kg/cm. 
To evenly distribute the load between the arm and the end-effector, the first joint of each arm incorporates two bus servos operating in tandem. 
Details on the design of the two end-effectors follow.

\subsubsection{Gripper End-effector Design}

The gripper end-effector is a two-DoF assembly that performs two motions: grasping the avocado and rotating its wrist to detach it from the peduncle and retrieve it. 
This harvesting procedure, along with a prototype end-effector, has been tested and validated in our previous work~\cite{zhou2024design}. 
That work informs the design of the gripper end-effector developed herein, with a major difference being the design optimization to perform the same task using a more lightweight assembly. Another key difference is that in the previous work, the wrist rotation was provided by the commercial robot arm, whereas herein this joint has been integrated into the developed end-effector. 

\begin{figure*}[ht]
\vspace{2pt}
    \centering
    \includegraphics[width=0.9\textwidth]{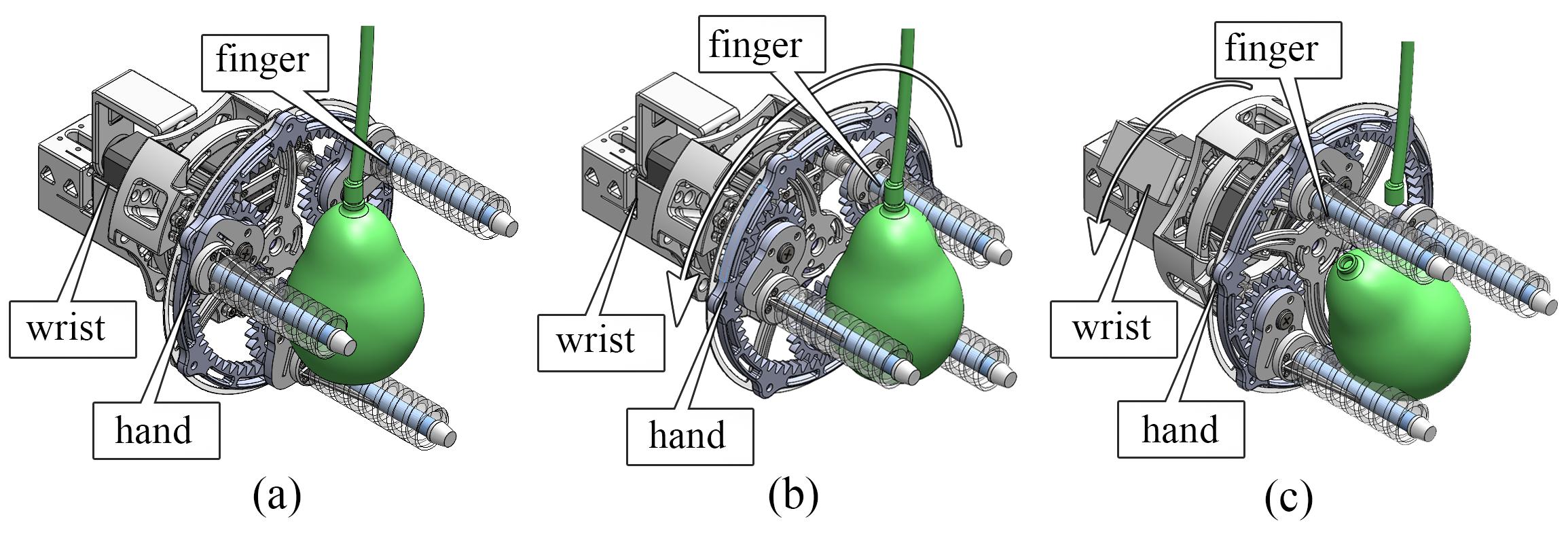}
    \vspace{-12pt}
    \caption{We employ three stages to harvest an avocado. (a) The arm brings the end-effector close to the avocado. (b) The base rotates to bring the fingers in contact with the avocado. (c) Once secured, the wrist rotates to apply a moment to the avocado to detach it from its peduncle and retrieve it.}
    \label{fig:3_stages_gripper}
    \vspace{-12pt}
\end{figure*}

The gripper end-effector contains three parts: the hand, fingers, and wrist. 
The hand is formed by one large internal spur gear which drives a small external spur gear located at the base of the fingers part. 
Each finger contains one slender cylinder covered by soft material, an iris plate part, and a small external gear activated passively by the hand's gear. 
The hand and the fingers together can be seen as an iris mechanism. 
The wrist employs one servo motor which is rigidly connected to the base of the hand and fingers.
The overall harvesting sequence is completed in three stages, depicted in Figure~\ref{fig:3_stages_gripper}. 
First (Figure~\ref{fig:3_stages_gripper}-(a)), the gripper arm moves the gripper end-effector to engulf the avocado within the hand. Then (Figure~\ref{fig:3_stages_gripper}-(b)), the hand of the gripper end-effector rotates counter-clockwise and brings the fingers in contact with the avocado. 
Note that the fingers are covered with soft material not only to prevent damage to the fruit but also to increase friction and contact robustness. 
Lastly (Figure~\ref{fig:3_stages_gripper}-(c)), the wrist of the end-effector starts to rotate once the avocado is secured within the fingers to detach it from its peduncle. 

\subsubsection{Fixer End-effector Design}
\begin{figure*}[ht]
\vspace{0pt}
    \centering
    \includegraphics[width=0.5\textwidth]{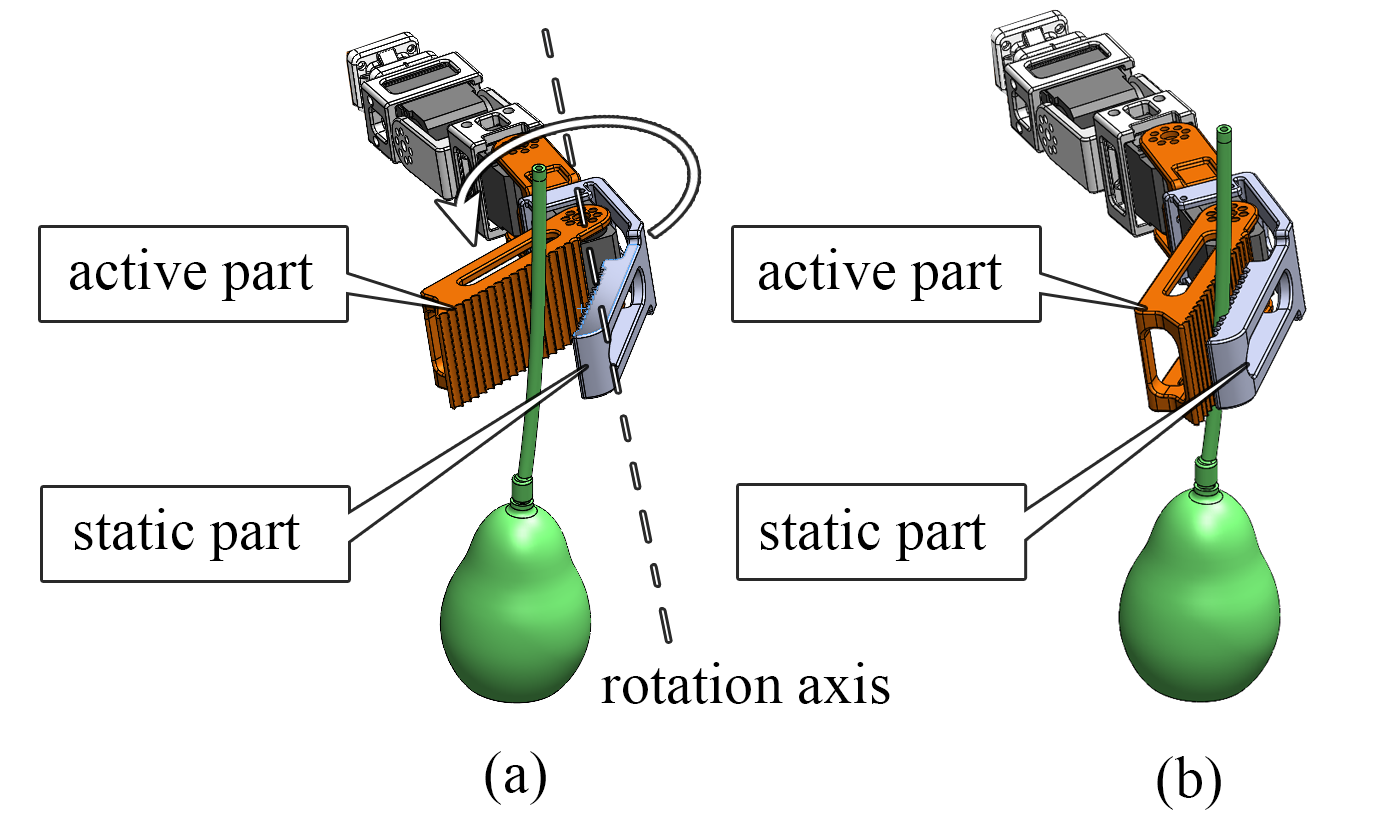}
    \vspace{-12pt}
    \caption{We employ two stages to secure the avocado peduncle. (a) The arm brings the end-effector close to the peduncle. (b) The active part rotates to come in contact with the peduncle.}
    \label{fig:fixer_workflow}
    \vspace{-12pt}
\end{figure*}

During experimentation in our previous work~\cite{zhou2024design}, we noticed that the peduncle of the avocado can store significant elastic energy which can lead to failed harvesting attempts via the three-stage process described previously. 
To address that limitation, we develop herein a fixer end-effector to hold the peduncle of the avocado tight before the three-stage process initiates. 
To this end, the fixer end-effector has only one DoF, containing one static part and one active part driven by another servomotor (Figure~\ref{fig:fixer_workflow}). 
The fixer arm first drives the fixer end-effector to a position where the peduncle is inside the two parts of the fixer end-effector (Figure~\ref{fig:fixer_workflow}-(a)). 
Then, the active part will start to rotate until the peduncle is grasped (Figure~\ref{fig:fixer_workflow}-(b)). 
As we will demonstrate later via physical experiments, this way can effectively keep the peduncle stationary while the moment is applied to the avocado to successfully harvest it. 

Importantly, for both end-effectors, two critical components of their overall operation involve (1) the detection of the avocado and peduncle, and (2) the dual-arm motion planning to bring the end-effectors to a pose that meets the presented first phase requirements.  
The development of these two components is presented in Section~\ref{sec:three} that follows. 
Before that, we derive the forward kinematics for each arm that is necessary for motion planning.

\subsection{Forward Kinematics}\label{sec:forward}
We use the Denavit-Hartenberg (DH) convention to derive the dual-arm system's forward kinematics. 
With reference to Figure~\ref{fig:dh_two_arm}, the DH parameters for the gripper and fixer arm are listed in Tables~\ref{table:dh_gripper_arm} and~\ref{table:dh_fixer_arm}, respectively. 
The subscript ``$g$'' is short for gripper whereas ``$f$" is short for fixer. 
Units for parameters $a_{ig}$, $a_{if}$, $d_{if}$ and $d_{ig}$ are in mm, whereas parameters $\alpha_{ig}$, $\alpha_{if}$, $\theta_{i}$ and $\phi_{i}$ are expressed in rad. 
Using the identified DH parameters, the forward kinematics homogeneous transformation for the gripper end-effector is given via~\eqref{eqn:T_gripper} while~\eqref{eqn:T_fixer} corresponds to the fixer end-effector forward kinematics. 
(Note that $c_{\theta_i}$ is short for $\cos \theta_i$, $s_{\theta_i}$ is short for $\sin \theta_i$, $i=1, 2, 3$, $c_{\phi_i}$ is short for $\cos \phi_i$, $s_{\phi_i}$ is short for $\sin \phi_i$, $i=1, 2, 3, 4$, $l=116.76$mm). 

From Figure~\ref{fig:dh_two_arm}, the base frame for the DH convention is noted as $x_{0}y_{0}z_{0}$ with its origin located at the midpoint of the line connecting fixer arm's first joint (origin of $x_{1f}y_{1f}z_{1f}$) and gripper arm's first joint (origin of $x_{1g}y_{1g}z_{1g}$.) 
Subscripts ``$ig$" (with $i=1, 2, 3, 4$) correspond to the gripper arm whereas subscripts ``$if$" (with $i=1, 2, 3, 4, 5$) is related to fixer arm. 
Note that in these derivations we have not considered the actual motion that each end-effector is performing; instead, the end-effector frames used in the forward kinematics analysis correspond to the point that is being used during dual-arm motion planning and control, to bring each end-effector to a desired configuration (the first phase of each case as described previously; this is often called the staging phase).  
Doing so considerably simplifies the kinematics analysis as well as the controller development, while offering a more direct way to match the visually-detected frame for a target avocado and its peduncle with the desired staging (i.e. first phase) pose for each end-effector.  

\begin{table}[ht]

\begin{center}
\caption{Gripper Arm DH Parameters.}
\vspace{-6pt}
\begin{tabular}{c c c c c} 
 \toprule
 Gripper link & $a_{ig}$ & $d_{ig}$ & $\alpha_{ig}$ & $\theta_i$\\
 \midrule
 link 1 & 142     & 0      & 0        & $\theta_1$ \\ 
 link 2 & 111     & 0      & -$\pi$/2 & $\theta_2$ \\ 
 link 3 & 200     & 15.975 & 0        & $\theta_3$ \\ 
 \bottomrule
\label{table:dh_gripper_arm}
\vspace{-12pt}
\end{tabular}
\end{center}
\hspace{0.15\textwidth}
\end{table}

\begin{table}[ht]
\begin{center}
\caption{Fixer Arm DH Parameters.}
\vspace{-6pt}
\begin{tabular}{c c c c c} 
 \toprule
 Fixer link & $a_{if}$ & $d_{if}$ & $\alpha_{if}$ & $\phi_i$\\
 \midrule
 link 1 & 142     & 0 & -$\pi$/2  & $\phi_1$ \\ 
 link 2 & 111     & 0 & $\pi$/2   & $\phi_2$ \\ 
 link 3 & 80      & 0 & -$\pi$/2  & $\phi_3$ \\ 
 link 4 & 122     & 0 & 0         & $\phi_4$ \\ 
 \bottomrule
\label{table:dh_fixer_arm}
\vspace{-12pt}
\end{tabular}
\end{center}
\end{table}

\begin{figure*}
    \begin{equation}
\label{eqn:T_gripper}
    T_{gripper} =
    \begin{bmatrix}
    1 & 0 & 0 & 0 \\
    0 & 1 & 0 & 0 \\
    0 & 0 & 1 & l \\
    0 & 0 & 0 & 1
    \end{bmatrix}
    \begin{bmatrix}
    c_{\theta_1} & -s_{\theta_1} & 0 & a_{1g} c_{\theta_1} \\
    s_{\theta_1} & c_{\theta_1}  & 0 & a_{1g} s_{\theta_1} \\
    0 & 0 & 1 & 0 \\
    0 & 0 & 0 & 1
    \end{bmatrix}
    \begin{bmatrix}
    c_{\theta_2} & 0 & -s_{\theta_2} & a_{2g} c_{\theta_2} \\
    s_{\theta_2} & 0 & c_{\theta_2}  & a_{2g} s_{\theta_2} \\
    0 & -1 & 0 & 0 \\
    0 &  0 & 0 & 1
    \end{bmatrix}
    \begin{bmatrix}
    c_{\theta_3} & -s_{\theta_3} & 0 & a_{3g} c_{\theta_3} \\
    s_{\theta_3} & c_{\theta_3}  & 0 & a_{3g} s_{\theta_3} \\
    0 & 0 & 1 & d_{3g} \\
    0 & 0 & 0 & 1
    \end{bmatrix}\;.
\end{equation}

\begin{equation}
\label{eqn:T_fixer}
\begin{split}
    T_{fixer} = &
    \begin{bmatrix}
    1 & 0 & 0 & 0 \\
    0 & 1 & 0 & 0 \\
    0 & 0 & 1 & -l \\
    0 & 0 & 0 & 1
    \end{bmatrix}
    \begin{bmatrix}
    c_{\phi_1} & 0 & -s_{\phi_1} & a_{1f} c_{\phi_1} \\
    s_{\phi_1} & 0 & c_{\phi_1}  & a_{1f} s_{\phi_1} \\
    0 & -1 & 0 & 0 \\
    0 & 0 & 0 & 1
    \end{bmatrix}
    \begin{bmatrix}
    c_{\phi_2} & 0 & s_{\phi_2}  & a_{2f} c_{\phi_2} \\
    s_{\phi_2} & 0 & -c_{\phi_2} & a_{2f} s_{\phi_2} \\
    0 & 1 & 0 & 0 \\
    0 & 0 & 0 & 1
    \end{bmatrix} \\
    &\begin{bmatrix}
    c_{\phi_3} & 0 & -s_{\phi_3} & a_{3f} c_{\phi_3} \\
    s_{\phi_3} & 0 & c_{\phi_3}  & a_{3f} s_{\phi_3} \\
    0 & -1 & 0 & 0 \\
    0 & 0 & 0 & 1
    \end{bmatrix}
    \begin{bmatrix}
    c_{\phi_4} & -s_{\phi_4} & 0 & a_{4f} c_{\phi_4} \\
    s_{\phi_4} & c_{\phi_4}  & 0 & a_{4f} s_{\phi_4} \\
    0 & 0 & 1 & 0 \\
    0 & 0 & 0 & 1
    \end{bmatrix}\;.
\end{split}
\end{equation}
\end{figure*}

\subsection{Workspace Analysis}\label{sec:workspace}

\begin{figure*}[ht]
    \centering
    \includegraphics[width=0.75\textwidth]{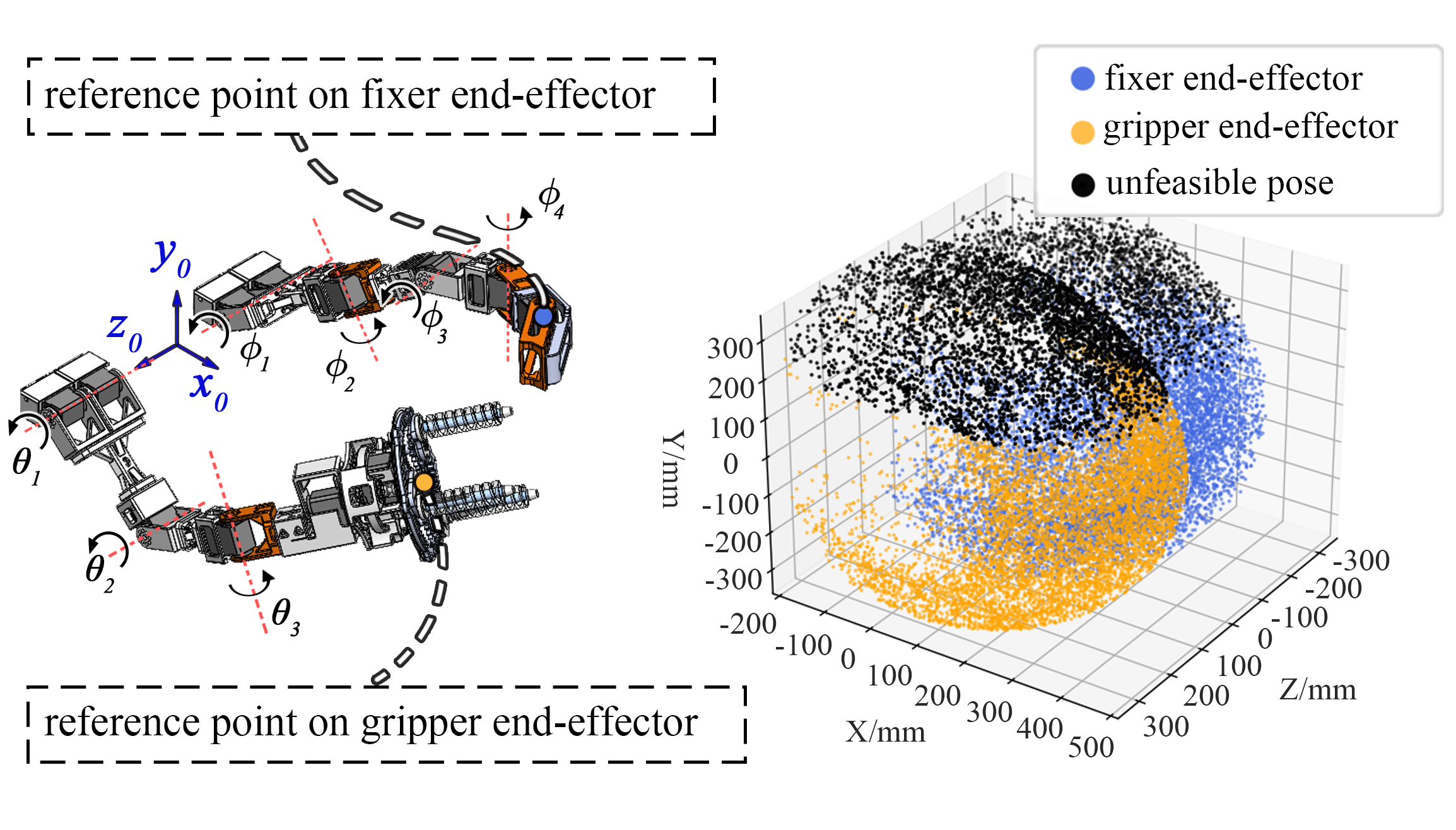}
    \vspace{-12pt}
    \caption{Workspace of the dual-arm system. (Figure best viewed in color.)}
    \label{fig:workspace}
    \vspace{-12pt}
\end{figure*}

Using the forward kinematics~\eqref{eqn:T_gripper} and~\eqref{eqn:T_fixer} and the physical joint limits we can identify the reachable configuration space for each arm. 
The joint limits for the gripper arm are $\theta_1\in[-\pi/2, \pi/2]$, $\theta_2\in[-\pi/2, \pi/2]$, and $\theta_3\in[-\pi/2, \pi/2]$. 
Similarly, the joint limits for the fixer arm are $\phi_1\in[-\pi/2, \pi/2]$, $\phi_2\in[-\pi/2, \pi/2]$, $\phi_3\in[-\pi/2, \pi/2]$ and $\phi_4\in[-\pi/2, \pi/2]$. 
The configuration spaces for both arms are visualized in Figure~\ref{fig:workspace}. 
For a harvesting task to be feasible, the avocado and (part of) its peduncle must both be within the intersection of the two end-effectors' workspaces. 
Configurations that lead to self-collision with the UAV chassis are denoted with black color in Figure~\ref{fig:workspace} and must be avoided. 

\section{Learning-based Visual Perception and Manipulation Planning for Autonomous Operation}
\label{sec:three}

\subsection{Autonomous Bimanual Aerial Robot Avocado Harvesting}

\begin{figure*}[ht]
\vspace{0pt}
    \centering
    \includegraphics[width=0.6\textwidth]{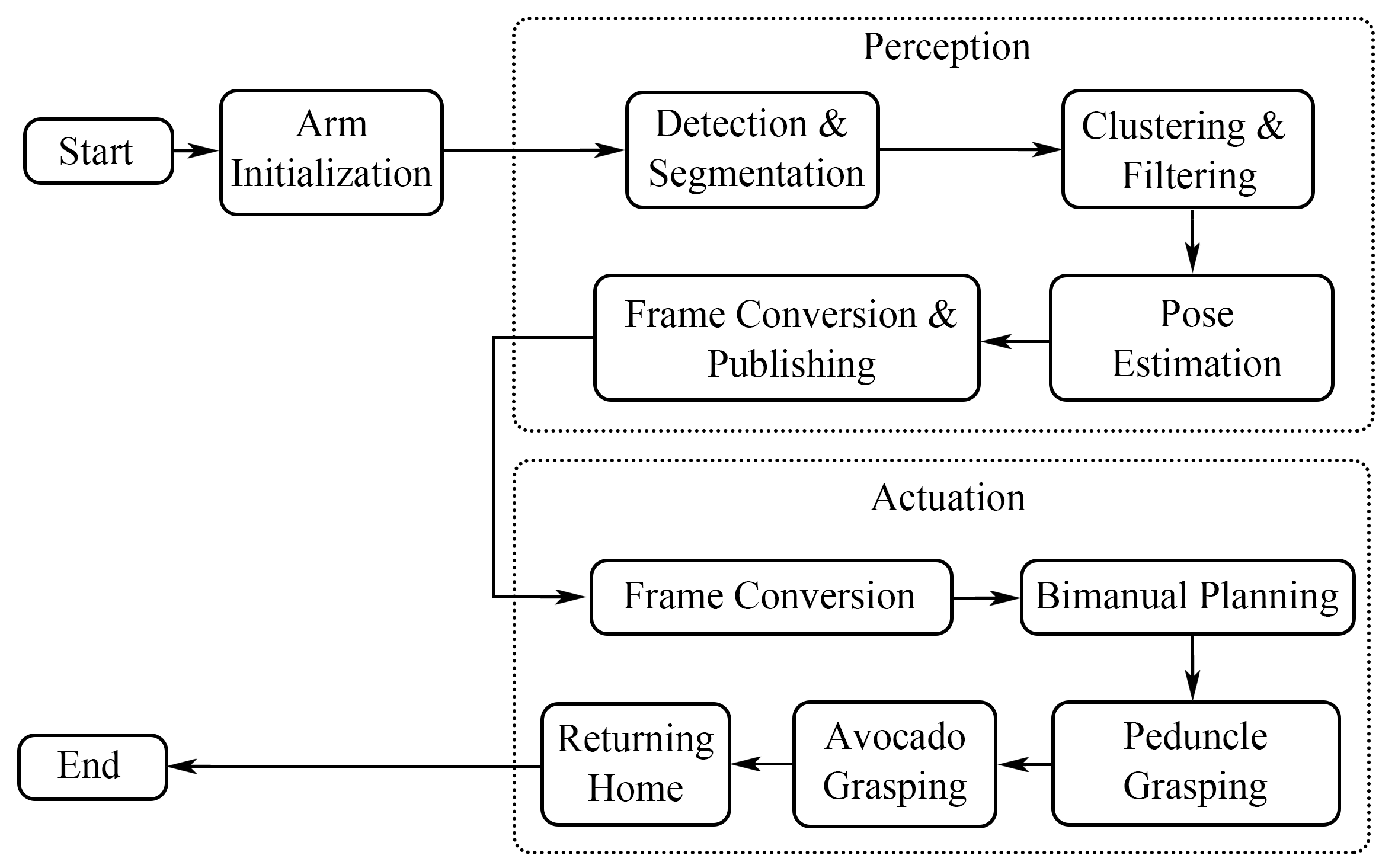}
    \vspace{-0pt}
    \caption{The overall framework deployed in this work to enable aerial bimanual manipulation for autonomous avocado harvesting.}
    \label{fig:real experiment flowchart}
    \vspace{-12pt}
\end{figure*}

The overall operational scheme is outlined in Figure~\ref{fig:real experiment flowchart}. 
There are two major components that we integrate. 
First, data from the depth camera are used to detect avocados and localize them in the scene. 
In earlier works~\cite{campbell2022integrated,dechemi2023robotic}, we used classical visual perception tools in a similar actuation-perception framework for leaf detection. 
In this work, we harness a visual-learning-based approach, discussed next. 
The perception module integrates detection and segmentation with clustering and filtering (Section~\ref{sec:31}) to estimate the pose (3D position and orientation) of a target avocado (Section~\ref{sec:32}). 
The inferred pose expressed in the camera frame is then transformed into the arm's coordinate system, which is the base frame in the DH analysis and the one used in bimanual manipulation planning. 

The camera and arm base frame are at a known and fixed offset distance expressed by vector $v_{Arm} = [0, 0.08653, 0.06436]^T$m in the arm base frame, and have a known and fixed orientation difference expressed by the rotation matrix 
\begin{figure*}
\begin{equation}
    \label{eqn:rotation_matrix}
    R_{Cam}^{Arm} = Rot_{x, 15^{\circ}} Rot_{z, 180^{\circ}} = 
    \begin{bmatrix}
        1 & 0                 & 0                  \\
        0 & \cos(15^{\circ}) & -\sin(15^{\circ}) \\
        0 & \sin(15^{\circ}) & \cos(15^{\circ})  \\                  
    \end{bmatrix}
    \begin{bmatrix}
        \cos(180^{\circ}) & -\sin(180^{\circ}) & 0 \\
        \sin(180^{\circ}) & \cos(180^{\circ})  & 0 \\
        0                 & 0                  & 1 \\
    \end{bmatrix}\;.
\end{equation}
\end{figure*}
Then, a point expressed in the camera frame, $P_{Cam}$ can be expressed into the arm base frame, $P_{Arm}$ via the homogeneous transformation 
\begin{equation}
    \label{eqn:axis conversion}
    P_{Arm} = v_{Arm} + R_{Cam}^{Arm} P_{Cam}\;.
\end{equation}

\begin{figure*}[ht]
\vspace{0pt}
    \centering
    \includegraphics[width=0.6\textwidth]{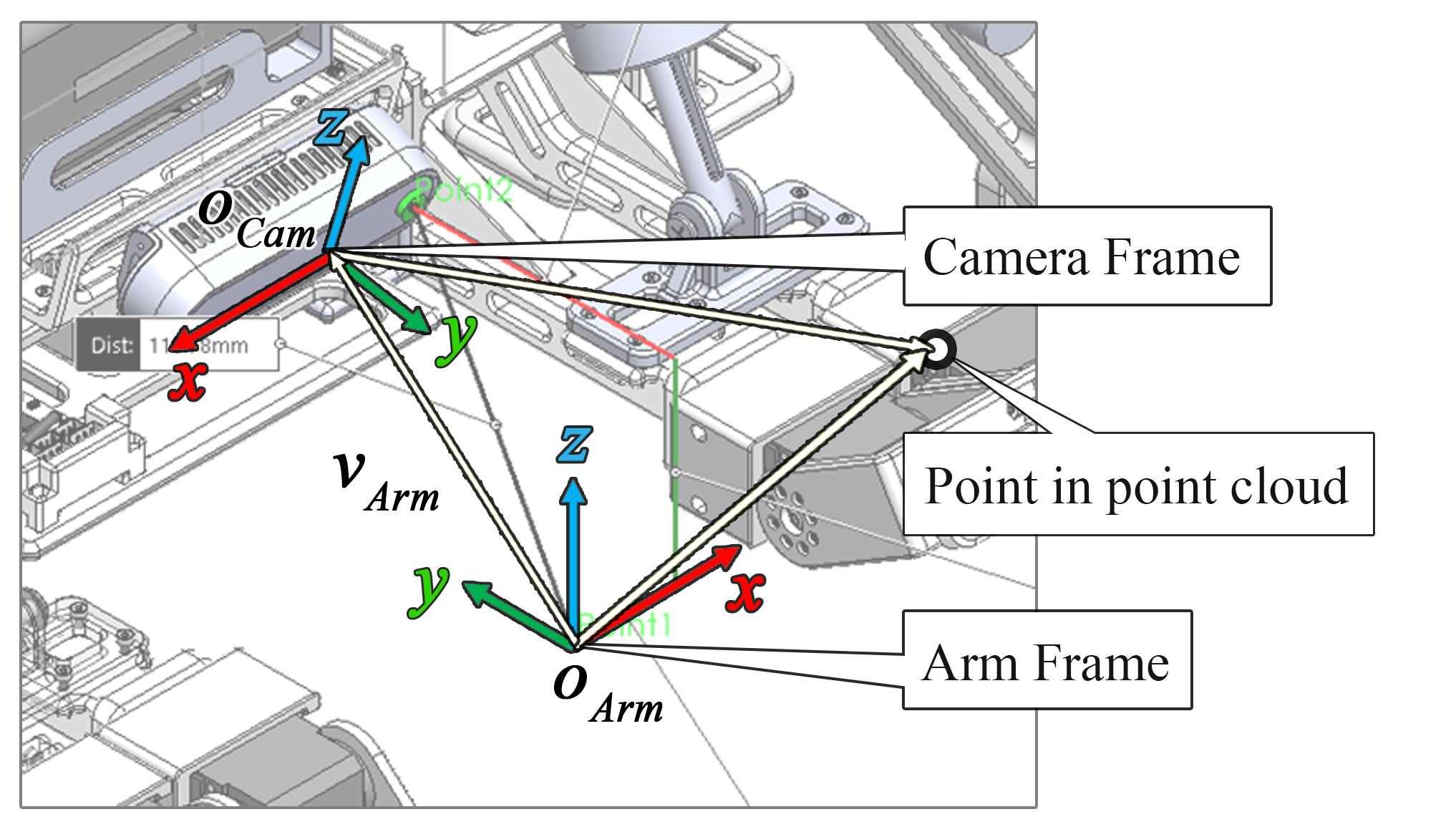}
    \caption{The fixed coordinate frame transformation between the camera frame and the dual-arm system's base frame.}
    \label{fig:axis_conversion}
    \vspace{-10pt}
\end{figure*}

During planning, two target poses are provided: one for the fixer arm to stage the fixed end-effector and then grasp the peduncle, and another for the gripper arm to stage its end-effector and grasp the avocado. 
Once this is completed, both arms return to their home configuration and the process can repeat. 
For motion planning, we use the MoveIt framework~\footnote{~https://moveit.ros.org/} by integrating the CAD file of our developed dual-arm system and carefully mapping physical joint limits and self-collision configurations (Section~\ref{sec:workspace}). 
If either target pose is outside the task-feasible subset of the workspace (i.e. the intersection of the two end-effectors' individual reachable workspaces as depicted in Figure~\ref{fig:workspace}), the UAV moves so that both target poses are within the task-feasible subset of the workspace and then hovers in place while the manipulation system turns active. 
All the different computational components running on different computing units are organized and communicate with each other via the Robot Operating System (ROS).\footnote{~https://ros.org/}

\subsection{Object Detection and Segmentation}\label{sec:31}

We use data from the stereo camera in a deep-learning-based visual perception framework to detect avocados in a scene and localize them with respect to the camera frame. 
Deep learning has been widely used in fruit detection due to its powerful ability to extract high-dimensional features from fruit images~\cite{koirala2019deep}.
Specifically, convolutional neural networks (CNNs) have demonstrated the potential to achieve accuracy and speed levels comparable to human performance in certain aspects of fruit detection. 
Among existing methods, YOLO has emerged as a popular choice. 
YOLO is a single-stage detector capable of performing both object identification and classification in a single shot, allowing the network to be smaller and faster~\cite{redmon2016you}. 
In this work, we use the YOLOv8~\cite{ultralytics2023yolov8} iteration owing to the improved performance in both object detection and segmentation.  

\begin{figure*}[ht]
\vspace{0pt}
    \centering
    \includegraphics[width=0.9\textwidth]{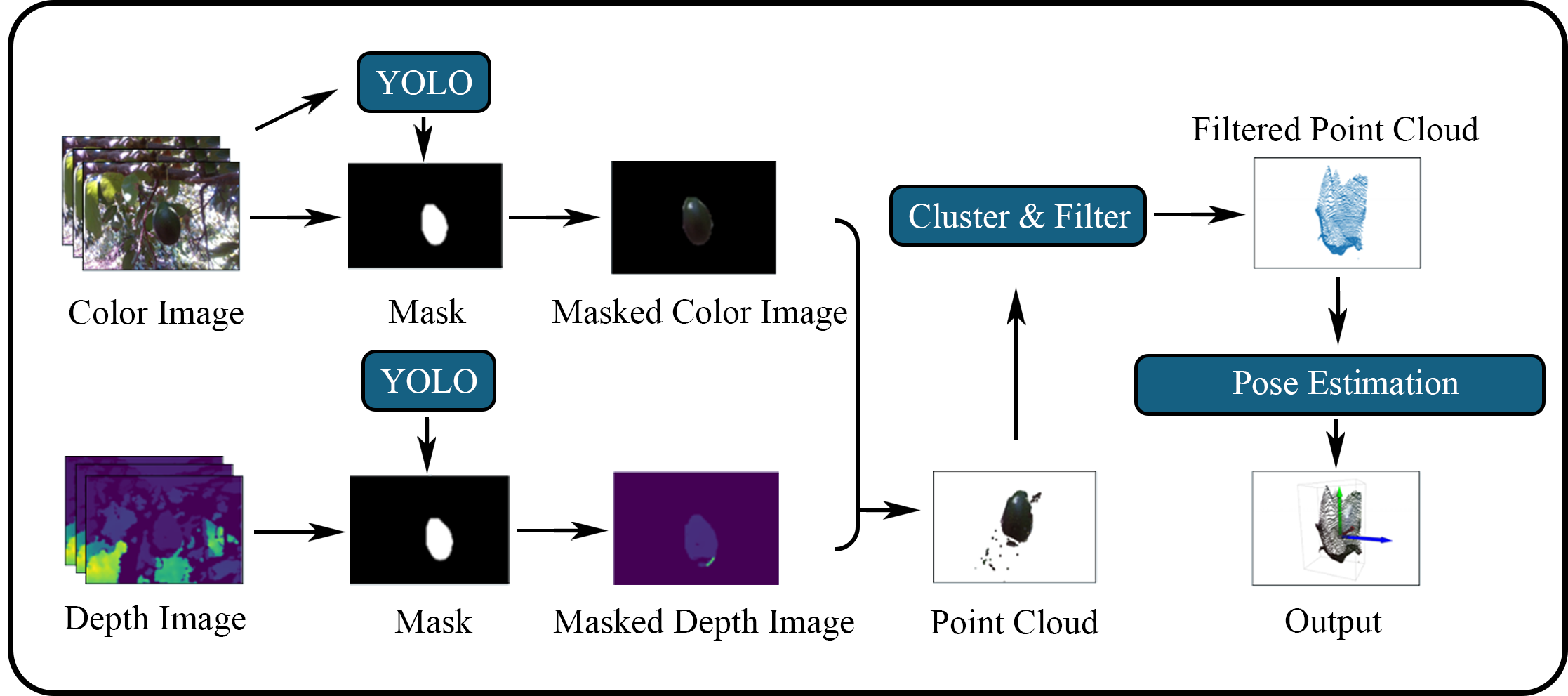}
    \vspace{-5pt}
    \caption{Our visual perception and learning pipeline for avocado detection and pose estimation.}
    \label{fig:visual}
    \vspace{-12pt}
\end{figure*}

Importantly, we seek to simultaneously extract the position in the image plane of a target avocado as well as infer its pose via the depth information. 
The main idea is that a 2D mask extracted from the color image is then mapped into the depth map to extract the corresponding pointcloud data (discussed in this section). 
The resulting masked depth images are then processed to infer the pose information of the avocado (discussed in the next section). 
Our developed visual perception and learning pipeline is summarized in Figure~\ref{fig:visual}.

Our project leverages a custom dataset~\cite{avocado-jvq5e_dataset} for finetuning the YOLOv8 framework to detect avocados in a given color image and segment out that area in the form of a 2D mask. 
The custom dataset comprises 1,072 meticulously annotated images, sourced from a combination of the Fruits-360 dataset~\cite{oltean2021fruits} and images collected in Agricultural Experiment Station (AES) avocado tree fields at the University of California, Riverside. 
To augment the dataset, we expanded the initial set of 1,072 images to a total of 2,580 images using various standardized augmentation techniques. 
This resulted in a training set of 2,262 images, a validation set of 211 images, and a test set of 107 images. 
The augmentations applied include horizontal flips, 0\% to 20\% zoom cropping, rotations ranging from $-15^\circ$  to $15^\circ$, and the addition of up to 0.1\% pixel noise. 
These augmentations were implemented to enhance the dataset's diversity, thereby improving the robustness and generalization capabilities of the trained models.

Fine-tuning different YOLOv8 networks allows for optimizing performance based on specific application requirements and computational constraints. 
The training was conducted using Ultralytics YOLOv8 on a system equipped with Python 3.9 and torch 2.3.0. 
The hardware setup included an NVIDIA GeForce RTX 3080 GPU with 10 MB of memory. 
A 300 epoch max buffer is set in all training instances; once concluded, the best results are reported. 
Table~\ref{tab:yolo_training} presents the training results of three YOLOv8 models: YOLOv8n-seg, YOLOv8s-seg, and YOLOv8m-seg. Each network varies in complexity, as indicated by the number of layers and parameters. 
YOLOv8n-seg, the smallest model, comprises 195 layers and 3,258,259 parameters, resulting in a compact size of 6.5 MB. 
This model has a recall of 0.9809 and a precision of 0.9695 after 166 epochs with a training time of 34.38 min. 
In contrast, YOLOv8s-seg, with the same number of layers but significantly more parameters (11,779,987), has a larger model size of 23.9 MB. This model achieves a recall of 0.9814 and a precision of 0.9843 after 240 epochs, with a total training time of 77.28 min. 
YOLOv8m-seg, the largest model with 245 layers and 27,222,963 parameters, and a size of 52.3 MB results in a recall of 0.9830 and a precision of 0.9845 after 131.40 min of training time over 208 epochs.

\begin{table*}[ht]
\caption{Training Results of YOLOv8 Models.}
\vspace{-6pt}
    \centering
    \begin{tabular}{c c c c c c c c}
         \toprule
        Networks & Layers & Parameters & Size (MB) & Epoch &  Training Time (M) & Recall & Precision \\
         \midrule
        YOLOv8n-seg & 195 & 3,258,259 & 6.5 & 166 & 34.38 & 0.9809 & 0.9695 \\
        YOLOv8s-seg & 195 & 11,779,987 & 23.9 & 240 & 77.28 & 0.9814 & 0.9843  \\
        YOLOv8m-seg & 245 & 27,222,963 & 52.3 & 208  & 131.40 & 0.9830 & 0.9845 \\
         \bottomrule
    \end{tabular}
    \label{tab:yolo_training}
\end{table*}

\begin{figure*}[ht]
    \centering
    \captionsetup{justification=centering}
    \includegraphics[width=0.95\textwidth]{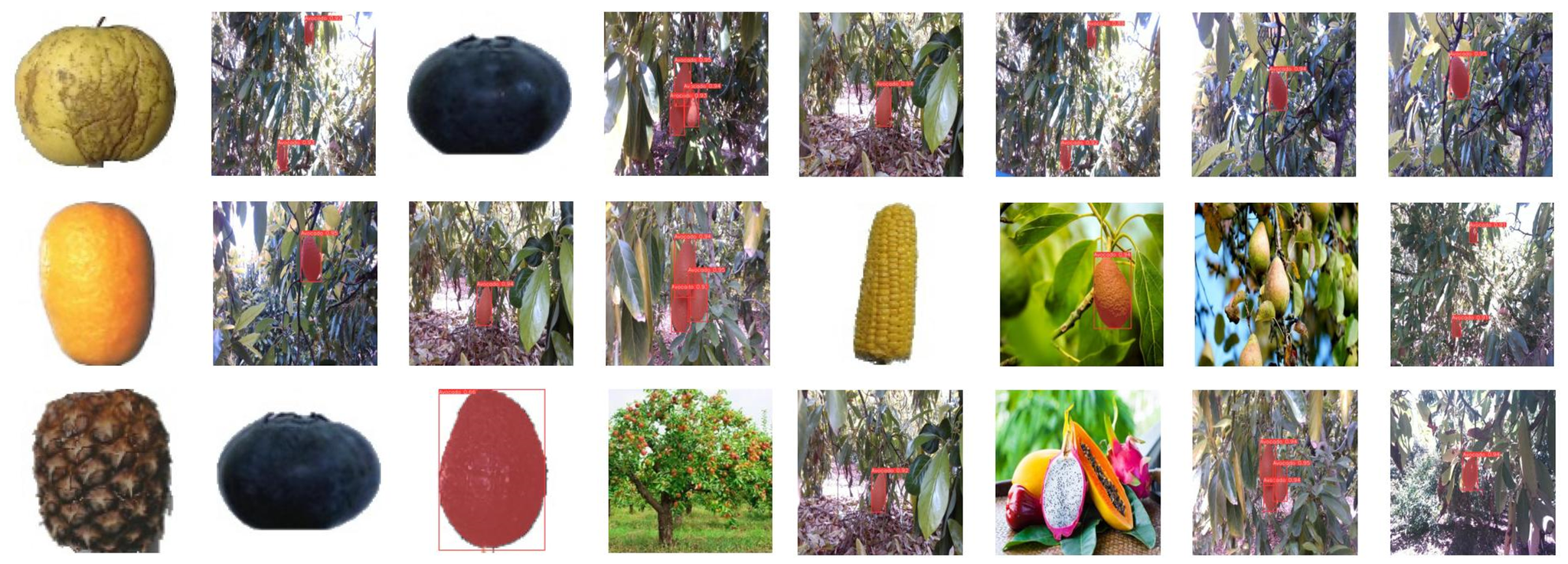}
    \vspace{-5pt}
    \caption{Inference Results of the Fine-Tuned YOLOv8n-seg Model on Test Set Images.}
    \label{fig:visual4}
    \vspace{-12pt}
\end{figure*}

Considering that the detector needs to run in real time, we also need to evaluate the inference time of different fine-tuned YOLO networks. 
We consider four computational units that can be integrated into a UAV: standard CPU (Intel NUC 10 with an i7-10710U processor), standard GPU (NVIDIA GeForce RTX 3080), NVIDIA Xavier NX board, and NVIDIA Orin Nano board. 
All configurations have a 100\% accuracy in detecting avocados in the test set images but come with different inference times (Table~\ref{tab:example}). 
As expected, the standard GPU is more capable in terms of real-time performance, achieving less than 6 ms inference time which is considered an optimal latency in real-time applications. 
However, it requires significantly more power which can be problematic for aerial robot deployment. 
The Xavier NX and Orin Nano achieve near-optimal real-time performance (i.e. slightly over 20 ms) for the smallest network, while the CPU is slower yet acceptable (less than 100 ms for the smallest network and slightly over for the medium-sized one). 
Based on this analysis, we select the YOLOv8n-seg model for the remainder of this study. Figure~\ref{fig:visual4} depicts some detection results in the test set images for this network. 
Also, considering the power requirements of the RTX GPU, we do not consider it any further. 

\begin{table*}[ht]
\caption{Inference Time [ms] on Various Platforms.}
\vspace{-6pt}
\centering
\begin{tabular}{c c c c c}
\toprule
Networks & CPU & RTX GPU & Xavier NX & Orin Nano\\
\midrule
YOLOv8n-seg & 48.06 & 0.92 & 29.40 & 24.79\\
YOLOv8s-seg & 111.07 & 2.15 & 46.89 & 46.59\\
YOLOv8m-seg & 226.44 & 4.38 & 91.27 & 77.09\\
\bottomrule
\end{tabular}
\label{tab:example}
\end{table*}

\subsection{Pose Estimation}\label{sec:32}
\begin{figure*}[ht]
\vspace{0pt}
    \centering
    \includegraphics[width=0.95\textwidth]{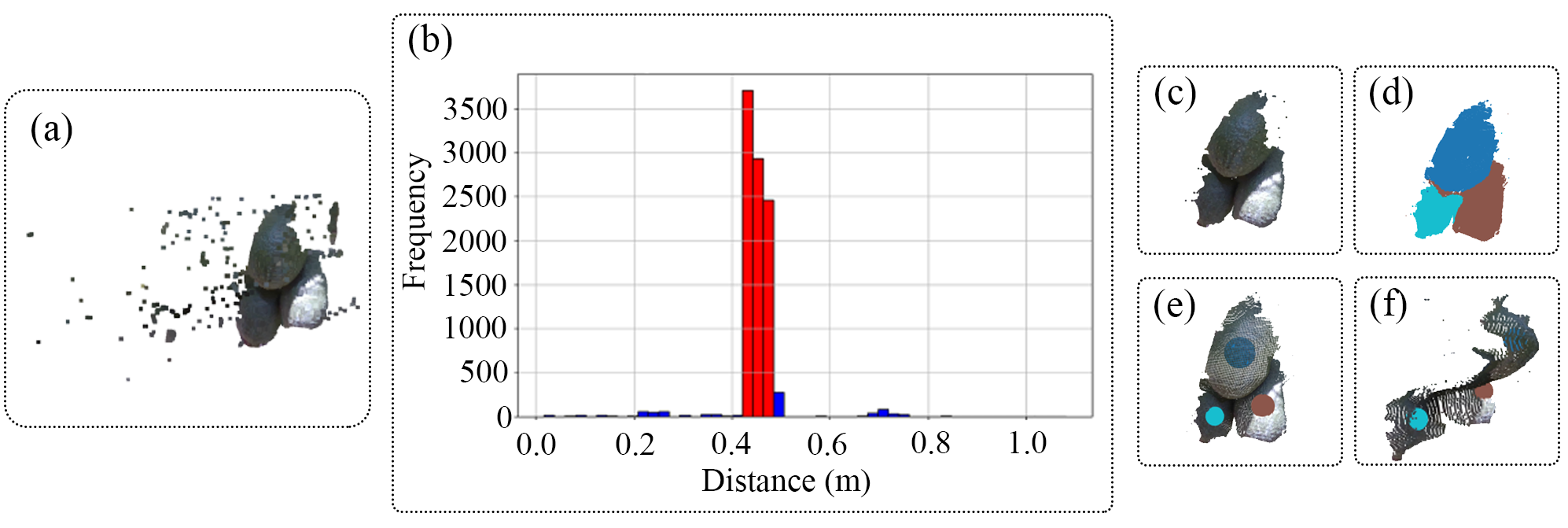}
    \vspace{-5pt}
    \caption{Point Clouds Processing by Histogram Filtering and Clustering.}
    \label{fig:visual2}
    \vspace{-12pt}
\end{figure*}

As illustrated in Figure~\ref{fig:visual}, the fine-tuned YOLOv8n-seg model detects and segments avocados in the color images (i.e. output a 2D mask). 
Given the different fields of view for the color and depth sensors of the camera, the depth images are preprocessed to match the color images. 
After applying the generated mask from the YOLO model to both the color and aligned depth images, 3D point clouds of the segmented avocados are created (Figure~\ref{fig:visual2}-(a)). 
These 3D point clouds are then used to localize suitable avocado candidates using 3D bounding boxes. 

As shown in Figure~\ref{fig:visual2}-(a), the point clouds may contain multiple candidates as well as random noise, therefore histogram filtering is applied. 
Histogram filtering is a technique used in image processing to improve the quality of data by removing noise or unwanted elements. The distribution of data values is analyzed (e.g., euclidean distances in this case) points falling outside a desired range are removed. 
We sort points according to their distance to the camera origin (Figure~\ref{fig:visual2}-(b)). 
The number of detections from the YOLO models is used as the ground truth for the number of clusters. 
In this case, three avocados are detected, and three bars are observed in the histogram (colored in red). 
The remaining 3D points after filtering out all those not belonging to the three selected bars are shown in Figure~\ref{fig:visual2}-(c). 
Clustering can also be performed using the histogram (Figure~\ref{fig:visual2}-(d)), where different colors represent different clusters (avocados in this case). 
Finally, the geometric center of the 3D points in each cluster is determined as the position of each detected avocado. 
Figures~\ref{fig:visual2}-(e) and (f) illustrate the estimated 3D positions of avocados from front and side views, respectively. 
It can be observed that when the point clouds capture relatively complete 3D shapes of the avocados, the estimated positions are closer to the true geometric centers of the fruits. 
However, when only a surface of 3D points is available, the estimated positions lie on the surfaces. 
Nonetheless, in each harvesting step, only the avocado candidate closest to the camera origin will be selected. 
After harvesting the first avocado, the depth camera will have better views of the remaining avocados, leading to a more accurate pose estimation of the remaining avocados.

\begin{figure*}[ht]
    \centering
    \includegraphics[width=0.7\textwidth]{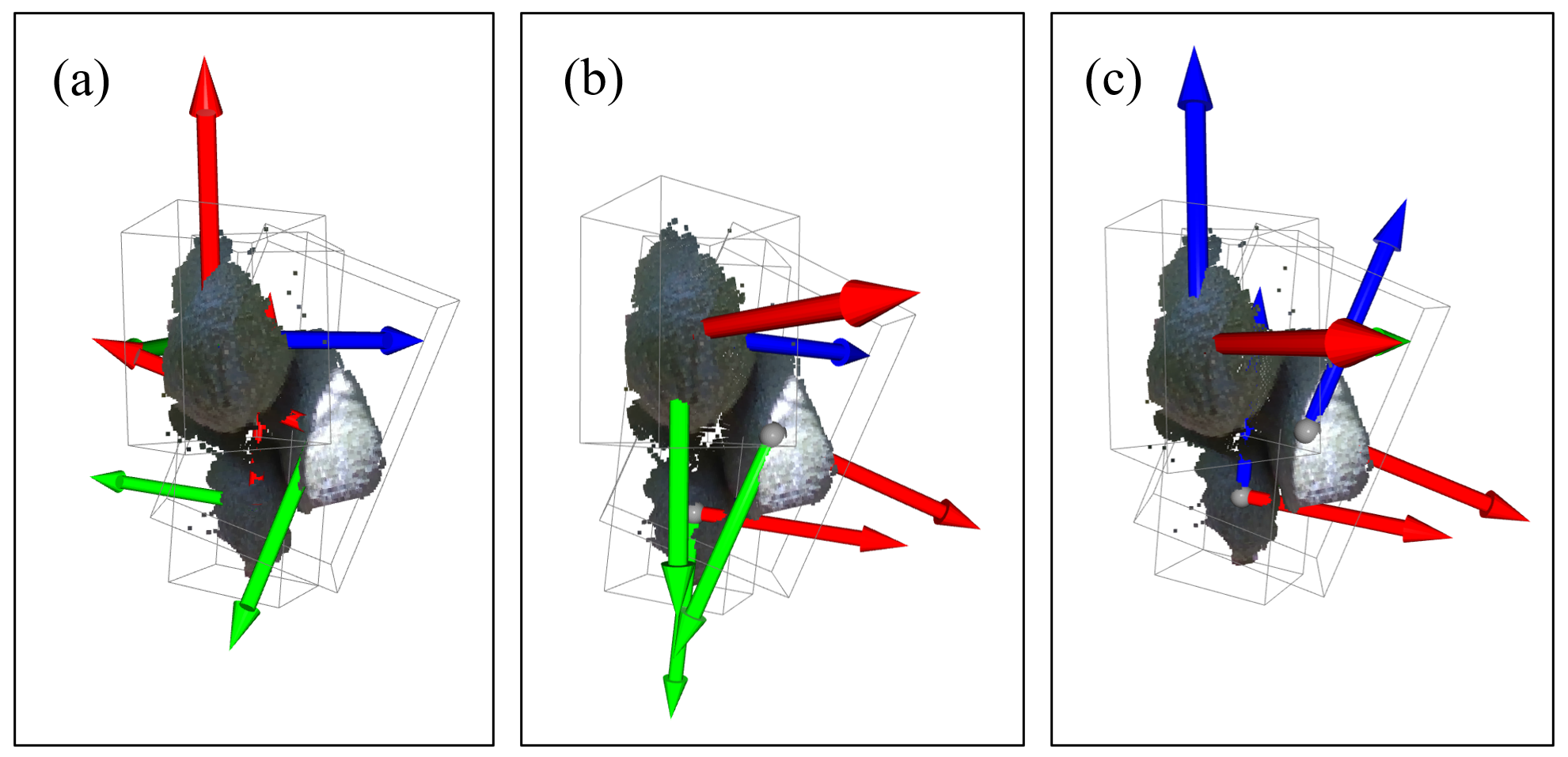}
    \vspace{-5pt}
    \caption{Avocado pose estimation and alignment correction of coordinate frames.}
    \label{fig:visual3}
    \vspace{-12pt}
\end{figure*}

To facilitate grasping, it is necessary to estimate both the position and the orientation of a target avocado. 
Pose estimation begins by drawing bounding boxes that cover the 3D points in each cluster. 
We adopt the oriented bounding box method in Open3D~\footnote{~https://www.open3d.org/} (Figure~\ref{fig:visual3}-(a)). 
However, it is often the case that the output coordinate frames may be assigned randomly even though the bounding boxes are fixed. 
To overcome this lack of deterministic behavior, we first post-process the generated coordinate systems by selecting the axis closest to the unit vector of the depth camera's coordinate system. 
The aligned coordinate systems are illustrated in Figure~\ref{fig:visual3}-(b), where the z-axes of the three detected avocados point downward, matching the z-axis of the camera frame. 
The refined coordinate systems are then converted to another frame, \( o_{Cam} \), as shown in Figure~\ref{fig:axis_conversion} so that the integration with the motion planner can be facilitated. 
We visualize the converted coordinate systems of the estimated orientations of all avocados in Figure~\ref{fig:visual3}-(c) and publish the Euler angles along with the 3D positions of the centers. 
The avocado candidate closest to the camera origin will be selected as the target for the aerial bimanual manipulator to reach. 

Lastly, we justify the selection of the onboard computing unit. 
Recall that the RTX GPU was excluded considering its higher power requirements compared to the CPU, Xavier NX, and Orin Nano. 
We now assess the total processing time of the perception module for the last three options. 
The fine-tuned YOLOv8n-seg is adopted in the perception framework. 
Based on our assessment (Table~\ref{tab:running_time}), pose estimation
requires an order of magnitude more computational time compared to detection and segmentation. 
Further, the CPU outperforms the two GPU-based boards in pose estimation by a large margin. 
Hence, we select the CPU option (the Intel NUC presented earlier) as the onboard computer for the robot to minimize the total running time. The perception framework runs at 1 Hz,  publishing the estimated position and orientation of the selected avocado candidate to the planner.

\begin{table}[ht]
\caption{Running Time [ms] for Perception Tasks on Various Platforms.}
\vspace{-6pt}
\centering
\begin{tabular}{c c c c}
\toprule
Step & CPU & Xavier NX & Orin Nano\\
\midrule
Detection \& Segmentation & 48.1 & 29.4 & 24.8\\
Pose Estimation & 268.2 & 1,491.0 & 496.2\\
Total & 316.3 & 1,520.4 & 521.0\\
\bottomrule
\end{tabular}
\label{tab:running_time}
\end{table}
\vspace{-12pt}

\section{Experimental Results and Discussion}
\label{sec:four}
\subsection{Preliminary Feasibility Experiment with the Dual-arm System}
\begin{figure*}[ht]
\vspace{0pt}
    \centering
    \includegraphics[width=0.9\textwidth]{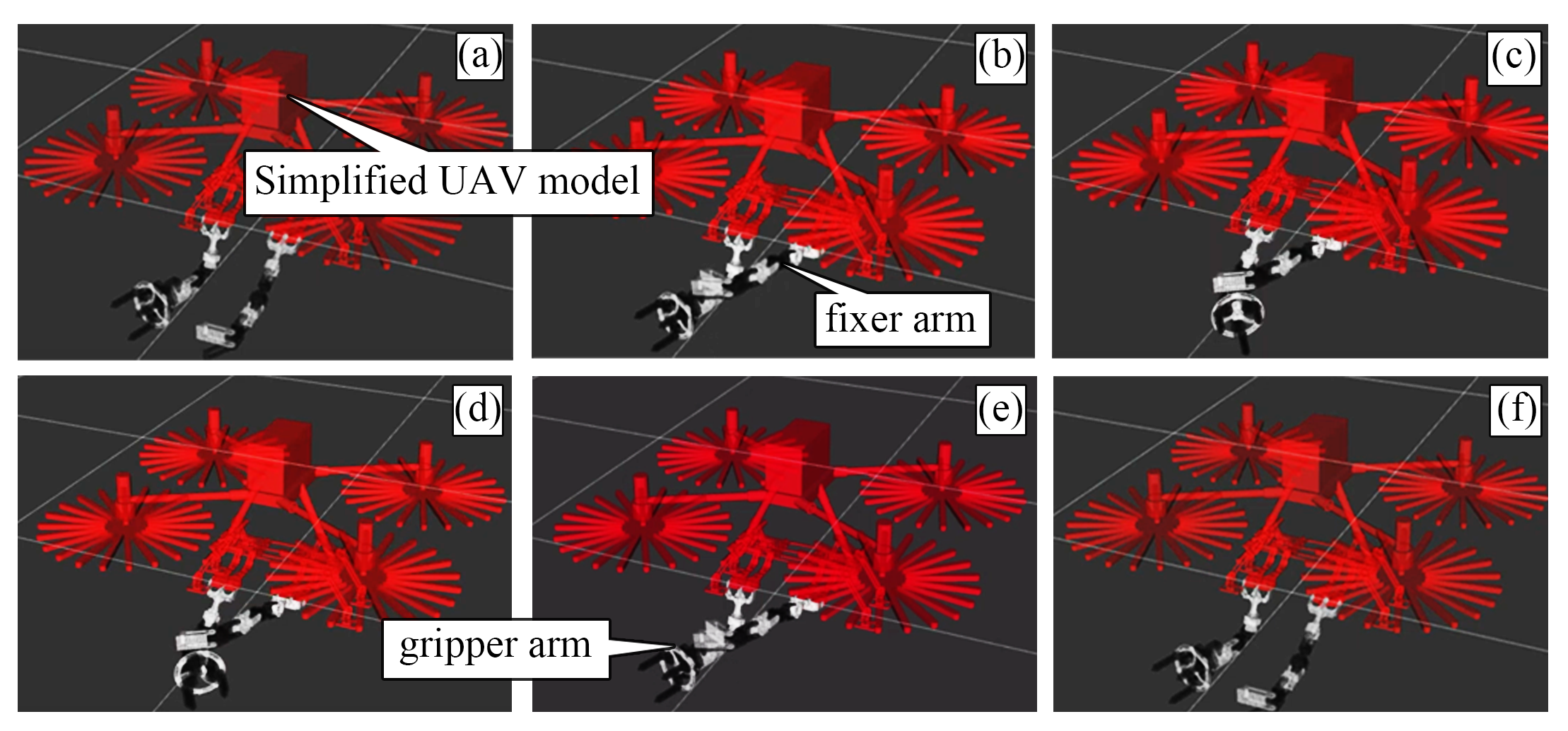}
    \vspace{-6pt}
    \caption{Visualization of different testing stages for the actions to be taken by each arm to harvest an avocado in simulation.}
    \label{fig:moveit preliminary experiment}
    \vspace{-12pt}
\end{figure*}

\begin{figure*}[ht]
\vspace{0pt}
    \centering
    \includegraphics[width=0.9\textwidth]{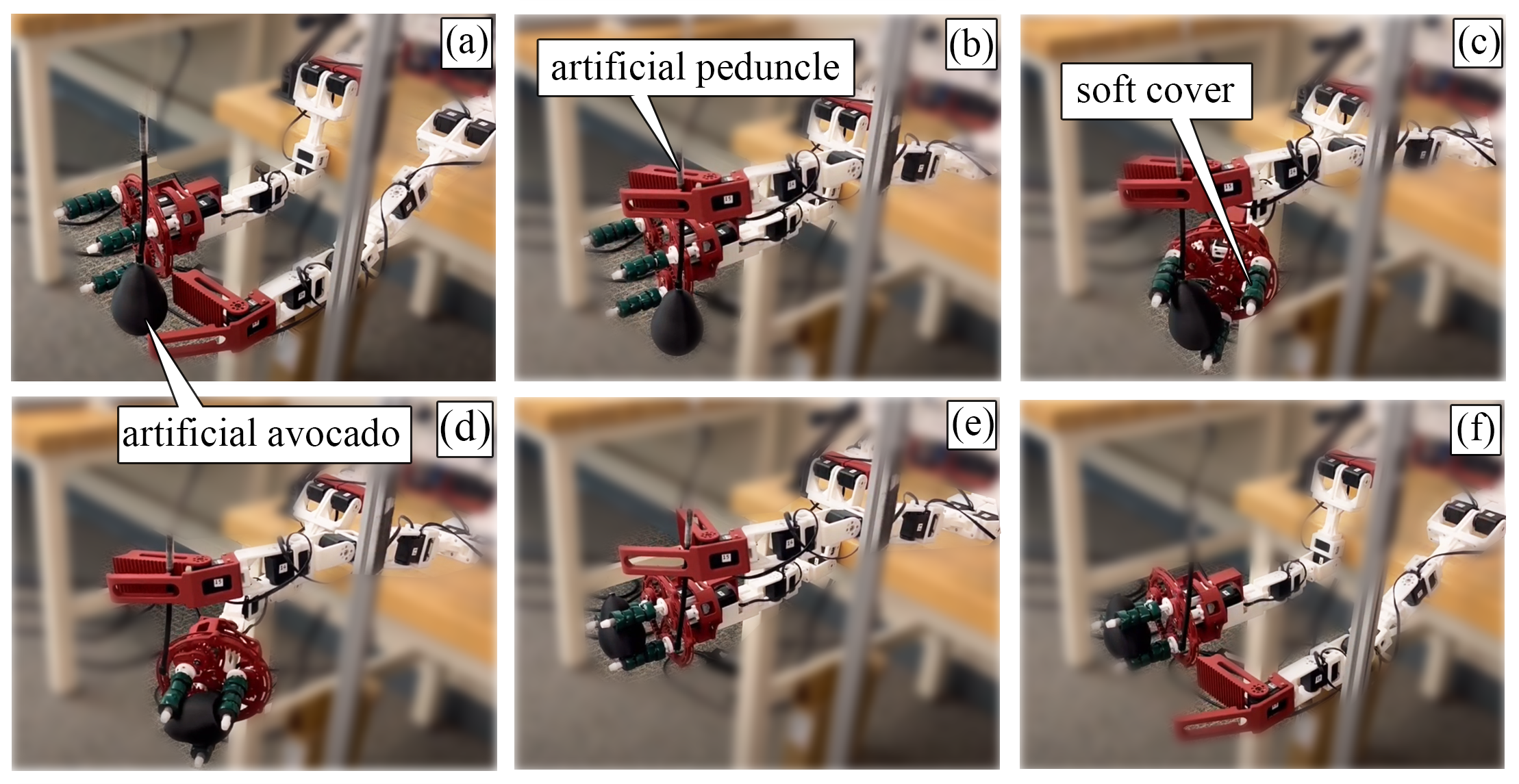}
    \vspace{-6pt}
    \caption{Harvesting an artificial avocado using the physical system in lab settings. (a) The dual-arm system starts at home configuration. (b) The fixer arm first drives the fixer end-effector to grasp the artificial peduncle. (c) The gripper arm drives the gripper end-effector to grasp the artificial avocado. (d) and (e) The gripper end-effector rotates its wrist to disengage the artificial avocado from the artificial peduncle. (f) Both arms return to their home position.}
    \label{fig:real preliminary experiment}
    \vspace{-12pt}
\end{figure*}

The purpose of the preliminary feasibility experiment is to demonstrate that the designed dual-arm system with the gripper and fixer end-effectors can work together to harvest an avocado. 
We conduct experiments both in simulation using a digital twin model we developed for our complete bimanual aerial robot (Figure~\ref{fig:moveit preliminary experiment}) as well as via physical experimentation in lab settings (Figure~\ref{fig:real preliminary experiment}). 
The hardware setup for the physical experimentation includes the dual-arm system mounted on the UAV, a standing steel frame, an artificial avocado with an artificial peduncle connected by magnets, and tape to fix the artificial peduncle on the standing steel frame. 
In this experiment, the UAV remains static and only the dual-arm system is active. 

Different stages of this experiment are indicated in panels (a)--(f) of Figures~\ref{fig:moveit preliminary experiment} and~\ref{fig:real preliminary experiment}; depicted states in simulation and hardware experiments match each other. 
The simulation showcases what actions are to be taken by each arm at different stages of avocado harvesting. 
Physical experiments are repeated 10 times to assess the success probability. Results are shown in Table~\ref{alttab:pre_experiment_results}.
Out of ten trials, the fixer arm has a 100\% success rate (i.e. securely grasping the artificial peduncle), while the gripper arm has an 80\% success rate (i.e. successfully retrieving the artificial avocado). 
Failure cases are primarily attributed to sub-optimal fixing points on the artificial peduncle, leading to the artificial avocado being in an unintended position. 
These initial tests confirm that the considered bimanual grasping procedure is feasible.

\begin{table}[htbp]
\centering
\caption{Preliminary Feasibility Experimental Results from 10 Trials Testing with an Artificial Avocado.}
\label{alttab:pre_experiment_results}
\vspace{-6pt}
\begin{tabular}{ccc}
\toprule
\textbf{ } & Success      &  Failure       \\ 
\midrule
\textbf{Gripper} & 8 & 2 \\ 
\textbf{Fixer} & 10 & 0 \\ 
\bottomrule
\end{tabular}
\end{table}

\subsection{Dual-arm System Testing in Field Experiments with Real Avocados}
\begin{figure*}[ht]
\vspace{-8pt}
    \centering
    \includegraphics[width=0.99\textwidth]{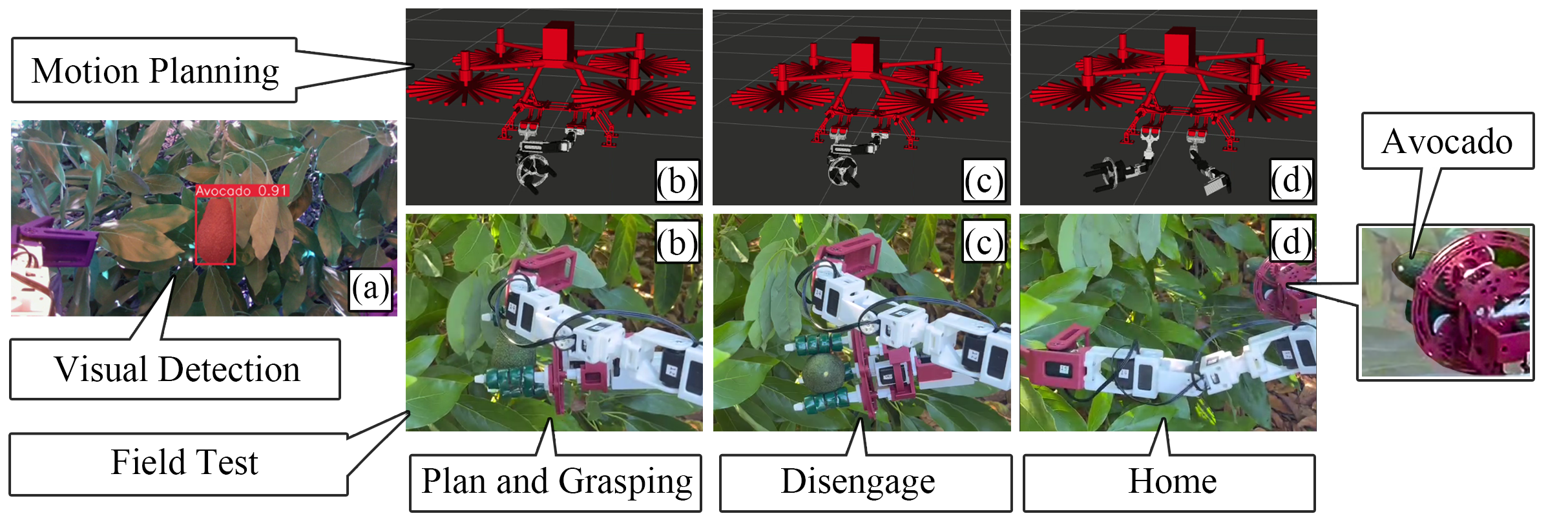}
    \vspace{-6pt}
    \caption{Field testing with the dual-arm system and visual detection: (a) Visual Detection. (b) Motion planning and bimanual grasping. (c) Gripper end-effector retrieving the avocado. (d) Fixer and gripper arms returning to their home configuration.} 
    \label{fig:field test}
    \vspace{-10pt}
\end{figure*}

In this experiment, the performance of visual detection and harvesting of real avocados in the orchard are being tested. 
Only the dual-arm system and camera are active in this set of experiments; the UAV is static. 
The main workflow of the experiment is shown in Figure~\ref{fig:field test}. 
The dual-arm system is first brought close to the tree so that target avocados are within the task-feasible subset of the reachable workspace. 
Then, the camera is activated to detect the avocado and return its estimated pose (Figure~\ref{fig:field test}-(a)). 
Once avocado pose information is broadcasted, the actuation module switches on to first transform from the camera coordinate frame to the arm frame and then call the planner to initiate bimanual planning, devising a trajectory for both the fixer arm and the gripper arm (Figure~\ref{fig:field test}-(b)). 
We are making the working assumption that the optimal point for peduncle grasping is situated $10$ cm above the avocado. 
Both arms bring their respective end-effectors at their staging pose. The fixed end-effector first grasps the peduncle. 
After a preset time (set here as 0.2 sec. regardless of the size of the avocado) the fingers on the gripper end-effector start to rotate and engulf the avocado. 
Once the avocado is grasped, the wrist joint on the gripper arm then rotates to detach the avocado (Figure~\ref{fig:field test}-(c)). After the avocado is collected, both arms return to their home position (Figure~\ref{fig:field test}-(d)). 
Overall, this experiment demonstrates that the developed detection and manipulation system can work in field settings. 

\subsection{Experimental Testing with the Complete Bimanual Aerial Manipulator}

\begin{figure*}[ht]
    \centering
    \includegraphics[width=0.99\textwidth]{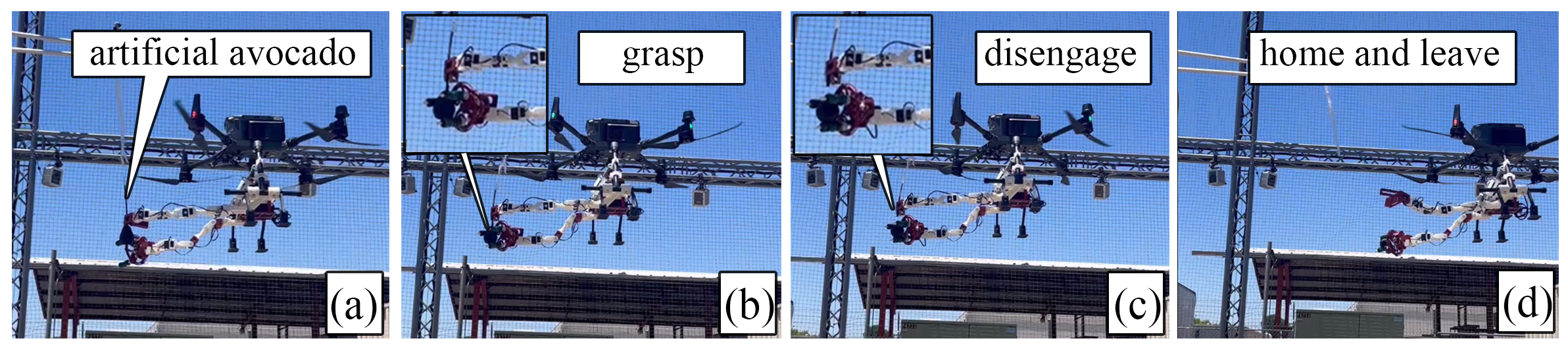}
    \vspace{-6pt}
    \caption{Complete system testing in controlled settings. (a) Fixer end-effector grasping the artificial peduncle. (b) Gripper end-effector grasping the artificial avocado. (c) The artificial avocado is being retrieved. (d) Fixer and gripper arms returning their home position, with the UAV leaving the designated position.}
    \label{fig:flying test}
    \vspace{-10pt}
\end{figure*}

The final experiment considers testing the complete system in controlled settings. 
The UAV is brought close to a target and then teleoperated to maintain safe control over its position and orientation. 
Additionally, an artificial avocado is used as the target object. 
The grasping procedure is the same as in the previous experiments. 
%
The experiment begins with the UAV flying to a designated position where the artificial avocado is located. 
Once the UAV reaches this position, the dual-arm sequence is initiated. 
The fixer end-effector is the first to engage, grasping the artificial peduncle (Figure \ref{fig:flying test}-(a)), and then the gripper end-effector moves into position to grasp the artificial avocado (Figure \ref{fig:flying test}-(b)) and retrieve it (Figure \ref{fig:flying test}-(c)). 
After harvesting the avocado, both arms return to their home configuration (Figure \ref{fig:flying test}-(d)), demonstrating that the arms can be safely stowed away after a task is completed to minimize any risk of collision or damage during the UAV's subsequent movements. 
In sum, this experiment showcases the practical application of the bimanual aerial robot in a controlled environment, providing valuable insights and demonstrating its potential in real-world scenarios of avocado harvesting. 

\section{Conclusion and Future Extension}
\label{sec:five}
Deploying robots for fruit harvesting holds the potential to improve efficiency and support sustainable farming practices. 
The use of aerial robots may be particularly appropriate in cases where the terrain and planting environment are unstructured, or the fruits to be harvested are in high-to-reach parts of a tree that would make a ground-robot-based solution financially and possibly technologically infeasible. 
However, aerial robots currently face significant challenges in performing such physical manipulation tasks autonomously. 
In this work, we focused on narrowing this gap by designing a bimanual aerial robot for avocado harvesting. 
Besides studying how to endow the physical capability to perform such tasks via aerial bimanual manipulation, we also endowed our system with integrated visual perception and learning capabilities to detect and estimate the pose of an avocado and plan its arms' motion to grasp it, all in real time while using only onboard resources available to the aerial robot. 
Several experiments help assess the efficacy of different developed components mostly in controlled lab settings but also in the field using both real and artificial avocados. 
Overall, results demonstrate the efficacy of the overall approach and support its potential for future field deployment. 

This work opens several avenues for future research. 
One important direction is to incorporate multi-stage perception and bimanual planning to enable continuous avocado harvesting. 
To achieve this, the perception module could leverage additional sensors, such as 3D LiDAR~\cite{gene2020lfuji} or thermal sensors~\cite{teng2023multimodal}. Simultaneously, the bimanual planning module could benefit from recent advancements in learning-based methods~\cite{grannen2023stabilize}. 
Another potential direction involves enhancing the resilience of the aerial robot to potential physical contacts~\cite{liu2021toward, liu2023contact, liu2023dynamic} that may occur when operating close to trees, thus facilitating their deployment in complex and challenging agricultural environments. 
Additionally, addressing the challenge of grasping avocados under extreme weather conditions, such as strong wind gusts, is crucial. 
Lastly, we aim to extend the application of the developed harvesting system to other crops besides avocados.

\section*{Acknowledgements}
The authors are with the Dept. of Electrical and Computer Engineering, University of California, Riverside. We gratefully acknowledge the support of NSF \# CMMI-2326309, USDA-NIFA \# 2021-67022-33453, and The University of California under grant UC-MRPI M21PR3417. Any opinions, findings, and conclusions or recommendations expressed in this material are those of the authors and do not necessarily reflect the views of the funding agencies.

\begingroup
\footnotesize  
\bibliographystyle{IEEEtran}
\bibliography{IEEEabrv, IEEEexample}
\endgroup

\end{document}